\documentclass[a4paper]{article}
\usepackage[utf8]{inputenc}
\usepackage[T1]{fontenc} 
\usepackage{RR,RRthemes}
\usepackage{hyperref}
\usepackage{lmodern}
\usepackage{textcomp}

\usepackage{microtype}
\usepackage{pgfplots}
\usepackage{bibspacing}
\usepackage{multirow}
\usepackage{array}
\usepackage{booktabs}
\usepackage{amsfonts}
\usepackage{paralist}
\usepackage{latexsym}
\usepackage{xcolor}
\usepackage{amsmath}
\usepackage{graphicx}

\renewcommand{\paragraph}[1]{\ \\[-0cm]\noindent\textbf{#1.}}

\newcommand{\ignore}[1]{}

\newcommand{\system}[1]{\textsc{#1}}
\newcommand{\paris}{\system{paris}}

\DeclareMathOperator{\type}{\mathit{type}}
\let\P\undef
\DeclareMathOperator{\P}{Pr}
\DeclareMathOperator{\E}{\mathbb{E}}

\renewcommand{\phi}{\varphi}
\renewcommand{\epsilon}{\varepsilon}

\RRdate{May 2011}
\RRversion{2}
\RRdater{August 2011}

\RRauthor{
Fabian M.~Suchanek\thanks{INRIA Saclay, 4 rue Monod, 91400 Orsay, France (fabian@{\textcolor{white}x}\hskip-2mm suchanek.name)}%
\and
Serge Abiteboul\thanks{LSV, 61, avenue du Pr\'esident Wilson, 94230 Cachan, France
  (serge.abiteboul@{\textcolor{white}x}\hskip-2mm inria.fr)}%
\and
Pierre Senellart\thanks{T\'el\'ecom ParisTech, 46 rue Barrault, 75013 Paris, France (pierre@{\textcolor{white}x}\hskip-2mm senellart.com)}%
}
\authorhead{Suchanek et al.}
\RRtitle{Alignement d'ontologies\\au niveau\\ des instances et des sch\'emas}
\RRetitle{Ontology Alignment\\at the Instance and Schema Level}
\titlehead{Suchanek et al. -- Ontology Alignment at the Instance and Schema Level}
\RRnote{This work has been partially funded by the
ERC grant Webdam.}
\RRresume{Nous pr\'esentons PARIS, un syst\`eme automatique d'alignement d'ontologies.
PARIS r\'ealise non seulement l'alignement d'instances de deux ontologies, mais aussi l'alignement des relations et des classes. L'alignement d'instances et de relations s'enrichissent
mutuellement. Notre approche, qui fournit ainsi une solution holistique au probl\`eme d'alignement d'ontologies, 
repose sur un mod\`ele probabiliste.
De ce fait, notre algorithme ne n\'ecessite ni param\`etre arbitraire ni r\'eglage manuel.
Nous d\'emontrons l'efficacit\'e de PARIS \`a travers des exp\'eriences sur des
ontologies diverses; nous obtenons des niveaux de pr\'ecisions
d'approximativement 90\% pour l'alignement de deux des plus
grandes ontologies publiquement disponibles.}
\RRabstract{We present PARIS, an approach for the automatic alignment of ontologies. PARIS aligns not only instances, but also relations and classes. Alignments at the instance-level cross-fertilize with alignments at the schema-level. Thereby, our system provides a truly holistic
solution to the problem of ontology alignment. The heart of the
approach is probabilistic. This allows PARIS to run without any parameter tuning. We demonstrate the efficiency of the
algorithm and its precision through extensive experiments. In
particular, we obtain a precision of around 90\% in experiments with two of the world's largest ontologies.}
\RRmotcle{Alignement, Ontologie, D\'esambiguation, RDFS,
Instances, Relations}
\RRkeyword{Alignment, Ontology, Disambiguation, RDFS,
Instances, Relations}
\RRprojet{Webdam}
\RRdomaine{2} 
\RCSaclay 

\begin{document}
 \RRNo{0408}

\makeRT 

\tableofcontents
\newpage

\section{Introduction}
\paragraph{Motivation}
An ontology is a formal collection of world knowledge. In this paper, we use the word \emph{ontology} in a very general sense, to mean both the schema (classes and relations), and the instances with their assertions. In recent
years, the success of Wikipedia and algorithmic advances in
information extraction have facilitated the automated construction of
large general-purpose ontologies. Notable endeavors of this kind
include \system{DBpedia}~\cite{dbpedia}, \system{KnowItAll}~\cite{knowitall},
\system{WikiTaxonomy}~\cite{Ponzetto07}, and \system{YAGO}~\cite{yago}, as well as
commercial services such as freebase.com, trueknowledge.com, and
wolframalpha.com. These ontologies are accompanied by a growing number
of knowledge bases\footnote{\url{http://www.w3.org/wiki/DataSetRDFDumps}}
in a wide variety of domains including:
music\footnote{\url{http://musicbrainz.org/}},
movies\footnote{\url{http://www.imdb.com/}}, geographical
data\footnote{\url{http://www.geonames.org/}},
publications\footnote{\url{http://www.informatik.uni-trier.de/~ley/db}},
medical and biological data\footnote{\url{http://www.uniprot.org/}}, or
government data\footnote{\url{http://www.govtrack.us/},
  \url{http://source.data.gov.uk/data/}}.

Many of these ontologies contain complementing data. For instance, a
general ontology may know who discovered a certain enzyme, whereas a
biological database may know its function and properties. However,
since the ontologies generally use different terms (identifiers) for
an entity, their information cannot be easily brought together. In
this respect, the ontologies by themselves can be seen as isolated
islands of knowledge. The goal of the Semantic Web vision is to interlink them, thereby creating one large body of universal
ontological knowledge \cite{lod2,lod}. This goal may be seen as a much
scaled-up version of record linking, with challenges coming from
different dimensions:
\begin{itemize}
\item unlike in record linkage, both instances and schemas should be
reconciled;
\item the semantics of the ontologies have to be respected;
\item the ontologies are typically quite large and complex. Moreover, we are interested in performing
the alignment in a fully automatic manner, and avoid tedious tuning or parameter settings.
\end{itemize}
A number of recent research have investigated this problem. There have been many works on entity
resolution (i.e., considering the A-Box only) \cite{ferrara2008,noessner2010,l2r,numprob,sigma,somemore,www2011eqv}.  In another direction,
much research has focused on schema alignment (i.e., considering the
T-Box only) \cite{gracia2009,jean2009,comapp,isaac2007,wang2008}.
However, in recent years, the landscape of ontologies
has changed dramatically. Today's ontologies often contain both a rich schema and, at the same time, a huge number of instances, with dozens of millions of assertions
about them. To fully harvest the mine of knowledge they provide,
their alignment has to be built on cross-fertilizing the alignments of
both instances and schemas.

In this paper, we propose a new, holistic algorithm for aligning 
ontologies. Our approach links not just related entity or
relationship instances, but also related classes and
relations, thereby capturing the fruitful interplay between schema
and instance matching. 
Our final aim is to discover and link
identical entities automatically across ontologies on a large scale,
thus allowing ontologies to truly complement each other.

\paragraph{Contribution}
The contribution of the present paper is three-fold:
\begin{itemize}

\item We present \paris{}\footnote{Probabilistic Alignment of Relations,
Instances, and Schema}, a probabilistic algorithm for aligning instances,
  classes, and relations simultaneously across ontologies.

\item We show how this algorithm can be implemented efficiently and
  that it does not require any tuning

\item We prove the validity of our approach through
  experiments on real-world ontologies.

\end{itemize}
The paper is organized as follows. Section~\ref{sec:relatedwork}
provides an overview of related work. We then 
introduce some preliminaries in Section~\ref{sec:prelim}. Section~\ref{sec:model} describes our
probabilistic algorithm and Section~\ref{sec:implementation} its
implementation. Section~\ref{sec:experiments} discusses experiments. To
ease the reading, some technical discussions are postponed to the appendix.

\section{Related Work}\label{sec:relatedwork}

\paragraph{Overview} The problem of ontology matching has its roots in the problem of identifying duplicate entities, which is also known as record linkage,
duplicate detection, or co-reference resolution. This problem has been
extensively studied in both database and natural language processing
areas \cite{datafusion,linkage-survey}. These approaches are less
applicable in the context of ontologies for two reasons. First, they do not consider the formal semantics that ontologies have (such as the \emph{subclassOf} taxonomy). Second, they focus on the alignment of instances and do not deal with the alignment of relations and classes.
There are a number of surveys and analyses that shed light on the problem of record linking in ontologies. Halpin et~al.~\cite{notsame} provide a good overview of the problem in general. They also study difficulties of existing \emph{sameAs}-links. These links are further analyzed by Ding et al.~\cite{sameasbeyond}. Glaser et al. ~\cite{glaser2009} propose a framework for the management of co-reference in the Semantic Web. Hu et al.~\cite{howmatchable} provide a study on how matches look in general. 

\paragraph{Schema Alignment} Traditional approaches to ontology matching have focused mostly either on aligning the classes (the ``T-Box'') or on matching instances (the ``A-Box''). The approaches that align the classes are manifold, using techniques such as sense clustering~\cite{gracia2009}, lexical and structural characteristics \cite{jean2009}, or composite approaches~\cite{comapp}. Unlike \paris{}, these approaches can only align classes and do not consider the alignment of relations and instances.
Most similar to our approach in this field are \cite{isaac2007} and
\cite{wang2008}, which derive class similarity from the similarities of
the instances. Both approaches consider only the equivalence of classes
and do not compute subclasses, as does \paris{}. Furthermore, neither can align relations or instances.

\paragraph{Instance Matching} There are numerous approaches to match
instances of one ontology to instances of another ontology. Ferrara,
Lorusso, and Montanelli~\cite{ferrara2008} introduce this problem from a
philosophical point of view. Different techniques are being used, such as
exploiting the terminological structure \cite{noessner2010}, logical
deduction \cite{l2r}, or a combination of logical and numerical methods
\cite{numprob}. The \system{Sig.ma} engine \cite{sigma} uses heuristics to match
instances. Perhaps closest to our approach is \cite{somemore}, which
introduces the concept of functionality. Different from their approach,
\paris{} does not require an additional smoothening factor.
The \system{silk} framework \cite{silk} allows specifying manual mapping rules.
The \system{ObjectCoref} approach by Hu, Chen, and Qu~\cite{www2011eqv} allows
learning a mapping between the instances from training data. With
\paris{}, we aim at an approach that uses neither manual input nor
training data. We compare some of the results of \system{ObjectCoref} to that of
\paris{} on the datasets of the
ontology alignment evaluation initiative~\cite{oaei2010} in
Section~\ref{sec:experiments}.
Hogan~\cite{Hogan2007} matches instances and propose to use these instances to compute the similarity between classes, but provides no experiments. Thus, none of these approaches can align classes and relations like \paris{}.

\paragraph{Holistic Approaches} Only very few approaches address the cause of aligning both
schema and instances: the \system{RiMOM} \cite{rimom} and \system{iliads} \cite{iliads}
systems. Both of these have only been tested on small ontologies. The
\system{RiMOM} system can align classes, but it cannot find \emph{subclassOf}
relationships. Furthermore, the approach provides a bundle of heuristics
and strategies to choose from, while \paris{} is monolithic. None of the
ontologies the \system{iliads} system has been tested on contained full-fledges
instances with properties. In contrast, \paris{} is shown to
perform well even on large-scale real-world ontologies with millions of instances.

\section{Preliminaries}\label{sec:prelim}

In this section, we recall the notions of ontology and of equivalence. Finally, we introduce the notion of functionality as one of the key concepts for ontology alignment.

\paragraph{Ontologies} We are concerned with ontologies available in the Resource Description
Framework Schema (RDFS~\cite{rdf}), the W3C standard for knowledge
representation. An RDFS ontology builds on \emph{resources}. A
resource is an identifier for a real-world object, such as a city, a
person, or a university, but also the concept of mathematics. For
example, \emph{London} is a resource that represents the city of
London. A \emph{literal} is a string, date or number. 
A \emph{property} (or \emph{relation}) is a binary
predicate that holds between two resources or between a resource and a
literal. For example, the property \emph{isLocatedIn} holds between
the resources \emph{London} and \emph{UK}. In the RDFS model, it is
assumed that there exists a fixed global set~$\mathcal{R}$ of resources, a fixed
global set $\mathcal{L}$ of literals, and a fixed global set $\mathcal{P}$ of
properties. Each resource is described by a URI. An RDFS
\emph{ontology} can be seen as a set of triples $O \subset \mathcal{R} \times \mathcal{P}
\times (\mathcal{R} \cup \mathcal{L})$, called \emph{statements}. In the
following, we assume given an ontology~$O$. To
say that $\langle x,r,y\rangle\in O$, we will write $r(x,y)$ and we
call $x$ and $y$ the \emph{arguments} of~$r$. Intuitively, such a
statement means that the relation $r$ holds between the entities $x$
and~$y$. We say that $x,y$ is a \emph{pair} of $r$. A relation
$r^{-1}$ is called the \emph{inverse} of a relation $r$ if $\forall
x,y: r(x,y) \Leftrightarrow r^{-1}(y,x)$. We assume that the ontology
contains all inverse relations and their corresponding
statements. Note that this results in allowing the first argument of a
statement to be a literal, a minor digression from the standard.

An RDFS ontology distinguishes between classes and instances. A class
is a resource that represents a set of objects, such as, e.g., the
class of all singers, the class of all cities or the class of all
books. A resource that is a member of a class is called an
\emph{instance} of that class. 
We assume that the ontology partitions
the resources into classes and instances.\footnote{RDFS
  allows classes to be instances of other classes, but in practice,
  this case is rare.} The \emph{rdf:type} relation connects an
instance to a class. For example, we can say that the resource
\emph{Elvis} is a member of the class of singers:
\emph{rdf:type(Elvis, singer)}.
A more specific class $c$ can be specified as a \emph{subclass} of a
more general class $d$ using the statement
\emph{rdfs:subclassOf(c,d)}. This means that, by inference, all
instances of $c$ are also instances of $d$. Likewise, a relation $r$
can be made a sub-relation of a relation $s$ by the statement
\emph{rdfs:subpropertyOf(r,s)}. This means that, by inference again,
$\forall x,y: r(x,y) \Rightarrow s(x,y)$. We assume that all such
inferences have been established and that the ontologies are available
in their \emph{deductive closure}, i.e., all statements implied by the
subclass and sub-property statements have been added to the ontology.

\paragraph{Equivalence}\label{phileqv}
In RDFS, the sets $\mathcal{P}$, $\mathcal{R}$, and $\mathcal{L}$ are global. That means that some
resources, literals, and relations may be \emph{identical} across
different ontologies. For example, two ontologies may contain the
resource \emph{London}, therefore share that resource. (In practice, \emph{London} is a URI, which makes it easy for two ontologies to use exactly the same identifier.) The semantics of RDFS
enforces that these two occurrences of the identifier refer to the same real-world object
(the city of London). The same applies to relations or literals that
are shared across ontologies. Conversely, two different resources can
refer to the same real-world object. For example, \emph{London} and
\emph{Londres} can both refer to the city of London. Such resources are
called \emph{equivalent}. We write
$\text{\emph{Londres}}\equiv\text{\emph{London}}$.

The same observation applies not just to instances, but also to
classes and relations. Two ontologies can talk about an identical
class or relation. They can also use different resources, but refer to
the very same real-world concepts. For example, one ontology can use
the relation \emph{wasBornIn} whereas another ontology can use the
relation \emph{birthPlace}. An important goal of our approach is to find out
that $\text{\emph{wasBornIn}}\equiv\text{\emph{birthPlace}}$.

In this paper, we make the following assumption: \emph{a given ontology does
  not contain equivalent resources.} That is, if an ontology contains
two instances $x$ and $x'$, then we assume $x\not\equiv x'$. We assume the same
for relations and classes. This is a reasonable assumption, because
most ontologies are either manually designed \cite{cyc,sumo}, or
generated from a database (such as the datasets mentioned in the
introduction), or designed with avoiding equivalent resources in mind
\cite{yago}. If the ontology does contain equivalent resources, then our
approach will still work. It will just not discover the equivalent resources 
within one ontology.
 
\paragraph{Functions}\label{sec:fun}
A relation $r$ is a \emph{function} if, for a given first argument,
there is only one second argument. For example, the relation
\emph{wasBornIn} is a function, because one person is born in exactly
one place. A relation is an \emph{inverse function} if its inverse is
a function. If $r$ is a function and if $r(x,y)$ in one ontology and
$r(x,y')$ in another ontology, then $y$ and $y'$ must be
equivalent. In the example: If a person is born in both \emph{Londres}
and \emph{London}, then $Londres\equiv London$. The same observation holds
for two first arguments of inverse functions.
As we shall see, functions play an essential role in deriving alignments
between ontologies.  Nevertheless, it turns out that the precise notion of
function is too strict for our setting. This is due to two reasons:
\begin{itemize}

\item First, a relation $r$ ceases to be a function as soon as there
  is one $x$ with $y$ and $y'$ such that $r(x,y)$ and $r(x,y')$. This means that
  just one erroneous fact can make a relation $r$ a
  non-function. Since real-world ontologies usually contain erroneous
  facts, the strict notion of function is not well-suited.

\item Second, even if a relation is not a function, it may contribute
  evidence that two entities are the same. For example, the relation
  \emph{livesIn} is not a function, because some people may live in
  several places. However, a wide majority of people live in one
  place, or in very few places.  So, if most people who live in
  \emph{London} also live in \emph{Londres}, this provides a strong
  evidence for the unification of \emph{London} and \emph{Londres}.
\end{itemize}
Thus, to derive alignments, we want to deal with ``quasi-functions''. This
motivates introducing the concept of
\emph{functionality}, as in \cite{somemore}. The \emph{local functionality} of a relation $r$
for a first argument $x$ is defined as 
\[
fun(r,x) = \frac{1}{\#y:r(x,y)}
\] 
where we write ``$\#y:\phi(y)$'' to mean
``$|\{y~|~\phi(y)\}|$''. Consider for example the relationship
$isCitizenOf$. For most first arguments, the functionality will be $1$,
because most people are citizens of exactly one country. However, for
people who have multiple nationalities, the functionality may be
$\frac{1}{2}$ or even smaller. The \emph{local inverse functionality}
is defined analogously as \[fun^{-1}(r,x) = fun(r^{-1},x)\]
 Deviating from~\cite{somemore}, we define
the \emph{global functionality} of a relation $r$ as the harmonic mean
of the local functionalities, which boils down to
\begin{equation}\label{globalfun} 
fun(r)= \frac{\#x: \exists y: r(x,y)}{\#x,y: r(x,y)}
\end{equation}
We discuss design alternatives and the rationale
of our choice in Appendix~%
\ref{app:globalfun}. The \emph{global inverse functionality} is
defined analogously as $fun^{-1}(r) = fun(r^{-1})$.

\section{Probabilistic Model}\label{sec:model}

\subsection{Equivalence of Instances}\label{sec:eqvinst}

We want to model the probability $P(x\equiv x')$ that one instance $x$ in
one ontology is equivalent to another instance $x'$ in another ontology.
Let us assume that both ontologies share a relation $r$. Following our
argument in Section~\ref{sec:fun}, we want the probability $P(x\equiv
x')$ to be large if $r$ is highly inverse functional, and if there are
$y\equiv y'$ with $r(x,y)$, $r(x',y')$ (if, say, $x$ and $x'$ share an
e-mail address). This can be written pseudo-formally as:

\[\exists r,y, y': r(x,y) \wedge r(x',y') \wedge y \equiv y' \wedge
fun^{-1}(r) \mbox{ is high } \Longrightarrow x \equiv x'
\]

\noindent We transform this logical rule into a probability assignment for $x\equiv
x'$, assuming independence and using the
formalization described in Appendix~\ref{app:problog}, obtaining thus:
\begin{equation}\label{p1} 
P_1(x\equiv x') := 1 - \prod_{r(x,y), r(x',y')}(
1-fun^{-1}(r)\times
P(y\equiv y'))
\end{equation}
In other words,
as soon as there is one relation $r$ with $fun^{-1}(r)=1$ and with
$r(x,y)$, $r(x',y')$, and $P(y\equiv y')=1$, it follows that $P_1(x\equiv x')=1$. We discuss a design alternative in Appendix~\ref{app:eqvalt}. 

Note that the probability of $x\equiv x'$ depends recursively on the
probabilities of other equivalences. These other equivalences may hold
either between instances or between literals. We discuss the probability
of equivalence between two literals in Section~\ref{sec:implementation}.
Obviously, we set $P(x\equiv x):=1$ for all literals and instances $x$.

Equation \eqref{p1} considers only positive evidence for an equality. To consider also evidence against an equality, we can use the following modification. We want the probability $P(x\equiv x')$ to be small, if there is a highly functional relation $r$ with $r(x,y)$ and if $y \not\equiv y'$ for all $y'$ with $r(x',y')$. Pseudo-formally, this can be written as 
\begin{equation}
\label{eq:punishment}
\exists r,y: r(x,y) \wedge (\forall
y': r(x',y') \Rightarrow y \not\equiv y')
\wedge fun(r) \mbox{ is high } \Longrightarrow x \not\equiv x.
\end{equation}
This can be modeled as 
\begin{equation}\label{p2} 
P_2(x\equiv x') := \prod_{r(x,y)} (1 - fun(r) \prod_{r(x',y')}
(1 -
P(y\equiv y')))
\end{equation}
As soon as there is one relation $r$ with $fun(r)=1$ and with $r(x,y)$,
$r(x',y')$, and $P(y\equiv y')=0$, it follows that $P_2(x\equiv x')=0$. We combine these two desiderata by multiplying the two probability estimates:
\begin{equation}\label{p3}
P_3(x\equiv x') := P_1(x\equiv x') \times P_2(x\equiv x')
\end{equation}
In the experiments, we found that Equation \eqref{p1} suffices in practice. However, we
discuss scenarios where Equation \eqref{p3} can be useful in Section~\ref{sec:experiments}.

\subsection{Subrelations}\label{sec:eqvrel}
The formulas we have just established estimate the equivalence between two entities that reside in two different ontologies, if there is a relation $r$ that is common to the ontologies. It is also a goal to discover whether a relation $r$ of one ontology is equivalent to a relation $r'$ of another ontology. More generally, we would like to find out whether $r$ is a sub-relation of $r'$, written $r \subseteq r'$.
Intuitively, the probability $P(r \subseteq r')$ is proportional to the number of pairs in~$r$ that are also pairs in $r'$:
\begin{equation}\label{prelnaive} 
P(r \subseteq r') := \frac{\#x,y : r(x,y) \wedge r'(x,y)}{\#x,y:
r(x,y) }
\end{equation}
The numerator should take into account the resources that have already
been matched across the ontologies. Therefore, the numerator is more
appropriately phrased as:

\[
\# x,y : r(x,y) \wedge (\exists x', y': x\equiv x' \wedge y\equiv y'
\wedge r'(x',y'))
\]
 
\noindent Using again our formalization from Appendix~\ref{app:problog}, this can
be modeled as:
\begin{equation}\label{prelnum}
\sum_{r(x,y)}(1-\prod_{r'(x',y')}(1-(P(x\equiv x')\times
P(y\equiv y'))))
\end{equation}
In the denominator, we want to normalize by the number of pairs in $r$ that have a counterpart in the other ontology. This is
\begin{equation}\label{prelden}
\sum_{r(x,y)}(1-\prod_{x',y'}(1-(P(x\equiv x')\times P(y\equiv
y'))))
\end{equation}
Thus, we estimate the final probability $P(r \subseteq r')$ as:
\begin{equation}
\frac{\sum_{r(x,y)}(1-\prod_{r'(x',y')}(1-(P(x\equiv
x')\times P(y\equiv
y'))))}{\sum_{r(x,y)}(1-\prod_{x',y'}(1-P(x\equiv
x')\times
P(y\equiv y')))} \label{prel}
\end{equation}
This probability depends on the probability that two instances (or
literals) are equivalent.

One might be tempted to set $P(r \subseteq r):=1$ for all relations $r$. However, in practice, we observe cases where the first ontology uses $r$ where the second ontology omits it. Therefore, we compute $P(r \subseteq r)$ as a contingent quantity.

We are now in a position to generalize Equation \eqref{p1} to the case
where the two ontologies do not share a common relation. For this, we
need to replace every occurrence of $r(x',y')$ by $r'(x',y')$ and factor
in the probabilities that $r' \subseteq r$ or $r\subseteq r'$.
This gives the following value to be assigned to $Pr(x\equiv x')$:
\begin{eqnarray}\notag
1 - \prod_{{r(x,y),r'(x',y')}} &( 1-P(r'
\subseteq r)\times fun^{-1}(r)\times
P(y \equiv y'))\\
\times &( 1-P(r
\subseteq r')\times fun^{-1}(r')\times
P(y \equiv y'))  \label{peqv1}
\end{eqnarray}
If we want to consider also negative evidence as in Equation~\eqref{p3}, we get
for $P(x\equiv x')$:
\begin{eqnarray}\notag
\Big(1 - \prod_{{r(x,y), r'(x',y')}} &(1-P(r'
\subseteq r)\times fun^{-1}(r)\times P(y\equiv y'))\\\notag
\times &( 1-P(r
\subseteq r')\times fun^{-1}(r')\times
P(y\equiv y'))\Big)\\\notag
\times \prod_{{r(x,y), 
r'}} & \Big(1 - fun(r) \times P(r'
\subseteq r) \times \prod_{r'(x',y')} (1 - P(x \equiv x'))\Big)\\
\times &\Big(1 - fun(r') \times P(r
\subseteq r') \times \prod_{r'(x',y')} (1 - P(x \equiv x'))\Big)\label{peqv2}
\end{eqnarray}

This formula looks asymmetric, because it considers only $P(r'
\subseteq r)$ and $fun(r)$ one one hand, and $P(r\subseteq r')$ and
$fun(r')$ on the other hand (and not, for instance, $P(r'
\subseteq r)$ together with $fun(r')$). Yet, it is not
asymmetric, because each instantiation of $r'$ will at some time also
appear as an instantiation of $r$. It is justified to consider $P(r'
\subseteq r)$, because a large $P(r' \subseteq r)$ implies that $r'(x,y)
\Rightarrow r(x,y)$. This means that a large $P(r' \subseteq r)$ implies
that $fun(r)<fun(r')$ and $fun^{-1}(r)<fun^{-1}(r')$.

If there is no $x',y'$ with $r'(x',y')$, we set as usual the last factor of the
formula to one, $\prod_{r'(x',y')} (1 - P(x \equiv x'))):=1$. This decreases $P(x\equiv x')$ in case one instance has relations that the other one does not have.
 
To each instance from the first ontology, our algorithm assigns multiple equivalent instances from the second ontology, each with a probability score. For each instance from the first ontology, we call the instance from the second ontology with the maximum score the \emph{maximal assignment}. If there are multiple instances with the maximum score, we break ties arbitrarily, so that every instance has at most one maximal assignment.
 
\subsection{Subclasses}\label{sec:eqvclass}

A class corresponds to a set of entities. One could be tempted to treat classes just like instances and compute their equivalence. However, the class structure of one ontology may be more fine-grained than the class structure of the other ontology. Therefore, we aim to find out not whether one class $c$ of one ontology is equivalent to another class $c'$ of another ontology, but whether $c$ is a subclass of $c'$, $c\subseteq c'$. Intuitively, the probability $P(c\subseteq c')$ shall be proportional to the number of instances of $c$ that are also instances of $c'$:
\[
P(c\subseteq c') = \frac{\# \; c\cap c'}{\# c}
\]. Again, we estimate the expected number of instances that are in both classes as
\[\E(\# \;c\cap c') =
\sum_{x:\type(x,c)}(1-\prod_{y:\type(y,d)}(1-P(x\equiv
y)))\] 
We divide this expected number by the total number of instances of $c$:
\begin{equation}
P(c \subseteq c') = 
\frac{\sum_{x:\type(x,c)}(1-\prod_{y:\type(y,d)}(1-P(x\equiv
y)))}
{\#x: \mathit{type}(x,c)}\label{pclass}
\end{equation}
The fact that two resources are instances of the same class can reinforce
our belief that the two resources are equivalent. Hence, it seems
tempting to feed the subclass-relationship back into Equation
\eqref{peqv1}. However, in practice, we found that the class information
is of less use for the equivalence of instances. This may be because of
different granularities in the class hierarchies. It might also be
because some ontologies use classes to express certain properties
(\emph{MaleSingers}), whereas others use relations for the same purpose
($\textit{gender}=\textit{male}$). Therefore, we compute the class equivalences only \emph{after} the instance equivalences have been computed. 

\section{Implementation}\label{sec:implementation}

\subsection{Iteration}

Our algorithm takes as input two ontologies. As already mentioned, we assume that a single ontology does not contain duplicate (equivalent) entities. This corresponds to some form of a domain-restricted unique name assumption. Therefore, our algorithm considers only equivalence between entities from different ontologies.
Strictly speaking, the functionality of a relation (Equation~\eqref{globalfun}) depends recursively on the equivalence of instances. If, e.g., every citizen lives in two countries, then the functionality of \emph{livesIn} is $\frac{1}{2}$. If our algorithm unifies the two countries, then the functionality of \emph{livesIn} jumps to 1. However, since we assume that there are no equivalent entities within one ontology, we compute the functionalities of the relations within each ontology upfront.

We implemented a fixpoint computation for Equations \eqref{prel} and
\eqref{peqv1}. First, we compute the probabilities of equivalences of
instances. Then, we compute the probabilities for sub-relationships. These
two steps are iterated until convergence. In a last step, the
equivalences between classes are computed by Equation \eqref{pclass} from
the final assignment. To bootstrap the algorithm in the very first step,
we set $P(r\subseteq r')=\theta$ for all pairs of relations $r,r'$ of
different ontologies. We chose $\theta=0.10$. The second round uses the
computed values for $P(r\subseteq r')$ and no longer $\theta$.
We have not yet succeeded in proving a theoretical condition under which
the iteration of Equations \eqref{prel} and \eqref{peqv1} reaches a
fixpoint. In practice, we iterate until the entity pairs under the
maximal assignments change no more (which is what we call convergence).
In our experiments, this state was always reached after a few iterations.
We note that one could always enforce convergence of such iterations by introducing a progressively increasing dampening factor.
Our model changes the probabilities of two resources being equal -- but
never the probability that a certain statement holds. All statements in
both ontologies remain valid. This is possible because an RDFS ontology
cannot be made inconsistent by equating resources, but this would not be
the case any more for richer ontology languages.

\subsection{Optimization}

The equivalence of instances (Equation \eqref{peqv1}) can be computed in
different ways. In the most naive setting, the equivalence is computed
for each pair of instances. This would result in a runtime of $O(n^2m)$,
where $n$ is the number of instances and $m$ is the average number of
statements in which an instance occurs (a typical value for $m$ is 20). This implementation took weeks to run one iteration. We overcame this difficulty as follows.

First, we optimize the computation of Equation \eqref{peqv1}. For each
instance $x$ in the first ontology, we traverse all statements $r(x,y)$
in which this instance appears as first argument. (Remember that we
assume that the ontology contains all inverse statements as well.) For
each statement $r(x,y)$, we consider the second argument $y$, and all
instances~$y'$ that the second argument is known to be equal to ($\{ y' :
P(y\equiv y')>0\}$). For each of these equivalent instances $y'$, we
consider again all statements $r(x',y')$ and update the equality of $x$
and $x'$. This results in a runtime of $O(nm^2e)$, where $e$ is the
average number of equivalent instances per instance (typically around
10). Equations \eqref{prel} and~\eqref{pclass} are optimized in a similar fashion.

Generally speaking, our model distinguishes \emph{true} equivalences ($P(x\equiv
x')>0$) from \emph{false} equivalences ($P(x\equiv x')=0$) and
\emph{unknown} equivalences ($P(x\equiv x')$ not yet computed). Unknown
quantities are simply omitted in the sums and products of the equations.
Interestingly, most equations contain a probability $P(x\equiv x')$ only
in the form $\prod (1-P(x\equiv x'))$. This means that the formula will evaluate to the same value if $P(x\equiv x')$ is unknown or if $P(x\equiv x')=0$. Therefore, our algorithm does not need to store equivalences of value $0$ at all.

Our implementation thresholds the probabilities and assumes every value
below $\theta$ to be zero. This greatly reduces the number of
equivalences that the algorithm needs to store. Furthermore, we limit the
number of pairs that are evaluated in Equations \eqref{prel} and \eqref{pclass} to $10,000$.
For each computation, our algorithm considers only the equalities of the
previous maximal assignment and ignores all other equalities. This
reduces the runtime by an order of magnitude without affecting much the
relation inclusion assessment. We stress that all these optimizations have for purpose to decrease the
running time of the algorithm \emph{without significantly
affecting the outcome of the
computation}. We have validated in our experiments that it is indeed the
case.

Our implementation is in Java, using the Java Tools developed for
\cite{leila} and Berkeley~DB. We used the Jena framework to load and
convert the ontologies. The algorithm turns out to be heavily IO-bound.
Therefore, we used a solid-state drive (SSD) with high read bandwidth
to store the ontologies. This brought the
computation time down from the order of days to the order of hours on
very large ontologies. We considered parallelizing the algorithm and
running it on a cluster, but it turned out to be unnecessary.

\subsection{Literal Equivalence}\label{liteqv}

The probability that two literals are equal is known a priori and will not change. Therefore, such probabilities can be set upfront (\emph{clamped}), for example as follows:
\begin{itemize}
\item The probability that two numeric values of the same dimension are equal can be a function of their proportional difference. 
\item The probability that two strings are equal can be inverse
proportional to their edit distance.
\item For other identifiers (social security numbers, etc.), the probability of equivalence can be a function that is robust to common misspellings. The checksum computations that are often defined for such identifiers can give a hint as to which misspellings are common.
\item By default, the probability of two literals being equal should be 0.
\end{itemize}
These functions can be designed depending on the application or on
the specific ontologies. They can, e.g., take into account unit conversions (e.g., between Kelvin and Celcius). They could also perform datatype conversions (e.g., between \emph{xsd:string} and \emph{xsd:anyURI}) if necessary. The probabilities can then be plugged into Equation \eqref{peqv1}. 

For our implementation, we chose a particularly simple equality function.
We normalize numeric values by removing all data type or dimension
information. Then we set the probability $P(x\equiv y)$ to $1$ if $x$
and $y$ are identical literals, to~$0$ otherwise. The goal of this work
is to show that even with such a simple, domain-agnostic, similarity
comparison between literals,
our probabilistic model is able to align ontologies with high precision;
obviously, precision could be raised even higher by implementing
more elaborate literal similarity functions.

\subsection{Parameters}

Our implementation uses the following parameters:
\begin{itemize}
\item The initial value $\theta$ for the equivalence of relations in the very first step of the algorithm. We show in the experiments that the choice of $\theta$ does not affect the results.
\item Similarity functions for literals. These are application-dependent.
However, we show that even with the simple identity function, the algorithm performs well.
\end{itemize} 
Therefore, we believe we can claim that our model has no dataset-dependent tuning parameters. Our algorithm can be (and in fact, was) run on all datasets without any dataset specific settings.
This contrasts \paris{} with other algorithms, which are often heavily dependent on parameters that have to be tuned for each particular application or dataset. Traditional schema alignment algorithms, for example, usually use heuristics on the names of classes and relations, whose tuning requires expertise  (e.g., \cite{rimom}). A major goal of the present work was to base the algorithm on probabilities and make it as independent as possible from the tuning of parameters. We are happy to report that this works beautifully. In order to improve results further, one can use smarter similarity functions, as discussed in Section~\ref{liteqv}.

\section{Experiments}\label{sec:experiments}
\subsection{Setup}
All experiments were run on a quad-core PC with 12 GB of RAM, running a
64bit version of Linux; 
all data was stored on a fast solid-state drive (SSD), with a
peak random access bandwidth of approximately 50~MB/s (to be compared
with a typical random access bandwidth of 1~MB/s for a magnetic hard
drive).

Our experiments always compute relation, class, and instance equivalences
between two given ontologies. Our algorithm was run until convergence
(i.e., until less than 1\,\% of the entities changed their maximal assignment). We
evaluate the instance equalities by comparing the computed final maximal
assignment to a gold standard, using the standard metrics of precision,
recall, and F-measure. For instances, we considered only the assignment
with the maximal score. For relation assignments, we performed a manual evaluation. Since \paris{} computes sub-relations, we evaluated the assignments in each direction. Class alignments were also evaluated manually. For all evaluations, we ignored the probability score that \paris{} assigned, except when noted.

\subsection{Benchmark Test}
To be comparable to \cite{www2011eqv,rimom,noessner2010,l2r}, we report
results on the benchmark provided by the 2010 edition of the ontology
alignment evaluation initiative (OAEI)~\cite{oaei2010}. We
ran experiments on two datasets, each of which consists of two
ontologies.\footnote{We could not run on the third dataset, because it
violates our assumption of non-equivalence within one ontology.} For each
dataset, the OAEI provides a gold standard list of instances of the
first ontology that are equivalent to instances of the second ontology.
The relations and classes are identical in the first and second ontology.
To make the task more challenging for \paris{}, we artificially renamed
the relations and classes in the first ontology, so that the sets of
instances, classes, and relations used in the first ontology are disjoint from the ones used in the second ontology.

For the person dataset, \paris{} converged after just 2 iterations and 2
minutes. For the restaurants, \paris{} took 3 iterations and 6 seconds.
Table~\ref{oaeires} shows our results.\footnote{Classes and relations
accumulated for both directions. Values for \system{ObjCoref} as reported
in \cite{www2011eqv}. Precision and recall are not reported in \cite{www2011eqv}. \system{ObjCoref} cannot match classes or relations.} We achieve near-perfect precision and recall, with the exception of recall in the second dataset. As reported in \cite{www2011eqv}, all other approaches \cite{rimom,noessner2010,l2r} remain below 80\,\% of F-measure for the second dataset, while only \system{ObjectCoref} \cite{www2011eqv} achieves an F-measure of 90\,\%. We achieve an F-measure of 91\,\%. We are very satisfied with this result, because unlike \system{ObjectCoref}, \paris{} does not require any training data. It should be further noted that, unlike all other approaches, \paris{} did not even know that the relations and classes were identical, but discovered the class and relation equivalences by herself in addition to the instance equivalences.

\begin{table}\tiny
\centering\begin{tabular}{llcrrrcrrrcrrr}
\toprule
&&\multicolumn{4}{c}{\textbf{Instances}} &
\multicolumn{4}{c}{\textbf{Classes}} &
\multicolumn{4}{c}{\textbf{Relations}}\\
\cmidrule(r){3-6}\cmidrule(lr){7-10}\cmidrule(l){11-14}
\textbf{Dataset}& \textbf{System}  &
\multicolumn{1}{c}{Gold} & \multicolumn{1}{c}{Prec} & \multicolumn{1}{c}{Rec} & \multicolumn{1}{c}{F}           & \multicolumn{1}{c}{Gold} & \multicolumn{1}{c}{Prec} & \multicolumn{1}{c}{Rec} & \multicolumn{1}{c}{F}         & \multicolumn{1}{c}{Gold} & \multicolumn{1}{c}{Prec} & \multicolumn{1}{c}{Rec} & \multicolumn{1}{c}{F}\\
\midrule
\multirow{2}{*}{\textbf{Person}} & \paris{}    & \multirow{2}{*}{500} &
100\,\% & 100\,\% & 100\,\%     & \multirow{2}{*}{4}    & 100\,\%& 100\,\%& 100\,\%
& \multirow{2}{*}{20}   &  100\,\% & 100\,\%   & 100\,\% \\
       & \system{ObjCoref} & & 100\,\% & 100\,\% & 100\,\%     &     & -    & -   & -         &     &  -   & -       & -     \\      
\midrule
\multirow{2}{*}{\textbf{Rest.}} & \paris{}    & \multirow{2}{*}{112} &
95\,\% & 88\,\%  & 91\,\%      & \multirow{2}{*}{4}    & 100\,\%& 100\,\%& 100\,\%
& \multirow{2}{*}{12}   &  100\,\% & 66\,\%   & 88\,\% \\
       & \system{ObjCoref} & & N/A      &  N/A     & 90\,\%      &     & -    & -   & -         &     & -     & -       & -     \\      
\bottomrule
\end{tabular}
\caption{Results (precision, recall, F-measure) of instance, class,
and relation alignment on OAEI datasets, compared with
\protect\system{ObjectCoref} \protect\cite{www2011eqv}. The ``Gold''
columns indicate
the number of equivalences in the gold standard.}
\label{oaeires}
\end{table}

\subsection{Design Alternatives}
To measure the influence of $\theta$ on our algorithm, we ran \paris{}
with $\theta=0.001,0.01,0.05,0.1,0.2$ on the restaurant dataset. A
larger $\theta$ causes larger probability scores in the first iteration.
However, the sub-relationship scores turn out to be the same, no matter
what value $\theta$ had. Therefore, the final probability scores are the
same, independently of $\theta$. In a second experiment, we allowed the
algorithm to take into account all probabilities from the previous
iteration (and not just those of the maximal assignment). This
changed the results only marginally (by one correctly matched entity), because the first iteration already has a very good
precision. In a third experiment, we allowed the algorithm to take into
account negative evidence (i.e., we used Equation \eqref{peqv2} instead
of Equation~\eqref{peqv1}). This made \paris{} give up all matches
between restaurants. The reason for this behavior turned out to be that
most entities have slightly different attribute values (e.g., a phone
number ``213/467-1108'' instead of ``213-467-1108''). Therefore, we
plugged in a different string equality measure. Our new measure
normalizes two strings by removing all non-alphanumeric characters and
lowercasing them. Then, the measure returns $1$ if the strings are equal
and $0$ otherwise. This increased precision to 100\,\%, but decreased recall to 70\,\%. Our experience with \system{YAGO} and \system{DBpedia} (see next experiment) indicates that negative evidence can be helpful to distinguish entities of different types (movies and songs) that share one value (the title). However, in our settings, positive evidence proved sufficient.
 
\subsection{Real-world Ontologies}
We wanted to test \paris{} on real-world ontologies of a large scale, with a rich class and
relation structure. At the same time, we wanted to restrict ourselves to
cases where an error-free ground truth is available. Therefore, we first chose
to align the \system{YAGO}~\cite{yago} and \system{DBpedia}
\cite{dbpedia} ontologies, and then to align \system{YAGO} with an
ontology built out of the \system{IMDb}\footnote{The Internet Movie Database, \url{http://www.imdb.com}}.

\begin{table}[h!]\small
\centering\begin{tabular}{lrrr}
\toprule
\textbf{Ontology}& \textbf{\#Instances} & \textbf{\#Classes} &
\textbf{\#Relations}\\
\midrule
\system{YAGO} & 2,795,289 & 292,206& 67\\
\system{DBpedia} & 2,365,777 & 318 &1,109\\
\system{IMDb} & 4,842,323 & 15 & 24 \\
\bottomrule
\end{tabular}
\caption{\system{YAGO} \protect\cite{yago}, \protect\system{DBpedia} \protect\cite{dbpedia} and \system{IMDb}.}\label{yagodbpedia}
\end{table}

\paragraph{\system{YAGO} vs.\ \system{DBpedia}}
 With several million instances, these are some
of the largest ontologies available. Each of them has thousands of
classes and at least dozens of relations. We took only the non-meta facts
from \system{YAGO}, and only the manually established ontology from
\system{DBpedia}, which yields the datasets described in Table
\ref{yagodbpedia}. Both ontologies use Wikipedia identifiers for
their instances, so that the ground truth for the instance matching can
be computed trivially.\footnote{We hid this knowledge from \paris{}.} However, the statements about the instances differ in both ontologies, so that the matching is not trivial. The class structure and the relationships of \system{YAGO} and \system{DBpedia} were designed completely independently, making their alignment a challenging endeavor.

We ran \paris{} for $4$ iterations, until convergence.
Table~\ref{ydres} shows the results per
iteration. To compute recall, we counted the number of shared instances in
\system{DBpedia} and \system{YAGO}. Since \system{YAGO} selects Wikipedia pages with many categories, and \system{DBpedia} selects pages with frequent infoboxes, the two resources share only 1.4 million entities. \paris{} can map them with a precision of 90\,\% and a recall of 73\,\%. If only entities with more than 10 facts in DBpedia are considered, precision and recall jump to 97\,\% and 85\,\%, respectively.
\begin{table}[h!]\tiny
\centering\begin{tabular}{rrrrrrrrrrrrrrr}
\toprule
\multicolumn{5}{c}{\bfseries Instances} & \multicolumn{5}{c}{\bfseries Classes} & \multicolumn{5}{c}{\bfseries Relations}\\
\cmidrule(r){1-5}\cmidrule(lr){6-10}\cmidrule(l){11-15}
Change & \multicolumn{1}{c}{Time} & \multicolumn{1}{c}{Prec} & \multicolumn{1}{c}{Rec} & \multicolumn{1}{c}{F} & 
\multicolumn{1}{c}{Time} & \multicolumn{2}{c}{\system{YAGO}$\:\subseteq\:$\system{DBp}}& \multicolumn{2}{c}{\system{DBp}$\:\subseteq\:$\system{YAGO}}& 
Time & \multicolumn{2}{c}{\system{YAGO}$\:\subseteq\:$\system{DBp}}& \multicolumn{2}{c}{\system{DBp}$\:\subseteq\:$\system{YAGO}}\\
to prev.    & & & &                                &      &
\multicolumn{1}{c}{Num}& \multicolumn{1}{c}{Prec}& \multicolumn{1}{c}{Num}& \multicolumn{1}{c}{Prec}& & \multicolumn{1}{c}{Num}& \multicolumn{1}{c}{Prec}& \multicolumn{1}{c}{Num}& \multicolumn{1}{c}{Prec}\\
\midrule
-    & 4h04m & 86\,\% & 69\,\% & 77\,\%       & -  & - & -       & - & -             & 19min  & 30 & 93\,\%& 134 & 90\,\%\\
12.4\,\% & 5h06m & 89\,\% & 73\,\% & 80\,\%       & -  & - & -       & - & -             & 21min  & 32 & 100\,\%& 144& 92\,\% \\
1.1\,\%  & 5h00m & 90\,\% & 73\,\% & 81\,\%       & -  & - & -       & - & -             & 21min  & 33 & 100\,\%& 149& 92\,\%\\
0.3\,\%& 5h26m & 90\,\% & 73\,\% & 81\,\%       & 2h14m & 137k & 94\,\% & 149 & 84\,\%     & 24min  & 33 & 100\,\%& 151& 92\,\%\\
\bottomrule
\end{tabular}
\caption{Results on matching \system{YAGO} and \system{DBpedia} over
iterations 1--4}\label{ydres}
\end{table}

\paris{} assigns one class of one ontology to multiple classes in the
taxonomy of the other ontology, taking into account the class inclusions.
Some classes are assigned to multiple leaf-classes as well. For our
evaluation, we excluded 19 high-level classes (such as
\emph{yagoGeoEntity}, \emph{physicalThing}, etc.). Then, we randomly sampled
from the remaining assignments and evaluated the precision manually. It
turns out that the precision increases substantially with the probability
score (see Figure~\ref{classprecplot}). We report the numbers for a
threshold of $0.4$ in Table~\ref{ydres} (the number of evaluated sample
assignments is 200 in both cases). The errors come from 3 sources: First,
\paris{} misclassifies a number of the instances, which worsens the precision
of the class assignment. Second, there are small  inconsistencies in the
ontologies themselves (\system{YAGO}, e.g., has several people classified
as \emph{lumber}, because they work in the wood industry). Last, there
may be biases in the instances that the ontologies talk about. For
example, \paris{} estimates that 12\,\% of the people convicted of murder
in Utah were soccer players. As the score increases, these assignments
get sorted out. Evaluating whether a class is always assigned to its most specific counterpart 
would require exhaustive annotation of candidate
inclusions. Therefore we only report the number of aligned classes and observe
that even with high probability scores (see Figure~\ref{classnumbplot}
and Table~\ref{ydres}) we find matches for a significant proportion
of the classes of each ontology into the other.
\begin{figure*}[h!]
\begin{minipage}{.48\linewidth}
\noindent\begin{tikzpicture}
\begin{axis}[xlabel=Threshold,ylabel={Precision},enlarge y
limits=0.05,enlarge x limits=false,width=\linewidth,extra x
ticks={0.1,0.9}]
\draw[very thin,gray] (axis cs:0.4,0) |- (axis cs:0,0.94);
\addplot[color=black,mark=x] table[x index=0,y index=2] {precision_classes.data};
\end{axis}
\end{tikzpicture}
\caption{\small Precision of class alignment
$\system{yago}\subseteq\system{DBpedia}$ as a function of the probability
threshold.}\label{classprecplot}
\end{minipage}\hspace{.04\linewidth}
\begin{minipage}{.48\linewidth}
\begin{tikzpicture}
\begin{axis}[xlabel=Threshold,ylabel={Number of classes ($\times
10{,}000$)},xmin=0.1,enlarge y
limits=0.05,enlarge x limits=false,width=\linewidth,extra x
ticks={0.1,0.9},scaled y ticks=real:10000,ytick scale label code/.code={}]
\addplot[color=black,mark=x] table[x index=0,y index=1] {count_classes.data};
\draw[very thin,gray] (axis cs:0.4,0) |- (axis cs:0,137254);
\end{axis}
\end{tikzpicture}
\caption{\small Number of \system{yago} classes that have at least one
assignment in \system{DBpedia} with a score greater than the
threshold.}\label{classnumbplot}
\end{minipage}
\end{figure*}

\begin{table}[h!]\tiny
\centering\begin{tabular}{l@{$\quad\subseteq\quad$}ll}
\toprule
\multicolumn{3}{c}{\system{YAGO}$\:\subseteq\:$\system{DBpedia}}\\
\midrule
y:actedIn	&  dbp:starring$^{-1}$ & 0.95\\
y:graduatedFrom	& dbp:almaMater&	0.93 \\
y:hasChild	& dbp:parent$^{-1}$&	0.53 \\
y:hasChild	& dbp:child	&0.30        \\
y:isMarriedTo	& dbp:spouse$^{-1}$	&0.56\\
y:isMarriedTo	& dbp:spouse	&0.89   \\
y:isCitizenOf	& dbp:birthPlace	&0.25\\
y:isCitizenOf	& dbp:nationality	&0.88\\
y:created	& dbp:artist$^{-1}$	&0.13\\
y:created	& dbp:author$^{-1}$	&0.17\\
y:created	& dbp:writer$^{-1}$	&0.30\\
\bottomrule
\end{tabular}\qquad
\begin{tabular}{l@{$\quad\subseteq\quad$}ll}
\toprule
 \multicolumn{3}{c}{\system{DBpedia}$\:\subseteq\:$\system{YAGO}}\\
\midrule
  dbp:birthName	& rdfs:label	&0.96\\
  dbp:placeOfBurial	& y:diedIn&	0.18\\
  dbp:headquarter	& y:isLocatedIn	&0.34\\
  dbp:largestSettlement	& y:isLocatedIn$^{-1}$	&	0.52\\
  dbp:notableStudent	& y:hasAdvisor$^{-1}$	&	0.10\\
  dbp:formerName	& rdfs:label	&0.73\\
  dbp:award	& y:hasWonPrize	&0.14\\
  dbp:majorShrine	& y:diedIn	&0.11\\
  dbp:slogan& 	y:hasMotto&	0.49\\
  dbp:author	& y:created$^{-1}$	&	0.70\\
  dbp:composer	& y:created$^{-1}$	&	0.61\\
\bottomrule
\end{tabular}\hspace*{-1em}
\caption{Some relation alignments between \system{YAGO} and \system{DBpedia} with their scores}\label{ydrel}
\end{table}
\begin{table}[h!]\tiny
\centering\begin{tabular}{rrrrrrrrrrrrrrr}
\toprule
\multicolumn{5}{c}{\bfseries Instances} & \multicolumn{5}{c}{\bfseries Classes} & \multicolumn{5}{c}{\bfseries Relations}\\
\cmidrule(r){1-5}\cmidrule(lr){6-10}\cmidrule(l){11-15}
Change & \multicolumn{1}{c}{Time} & \multicolumn{1}{c}{Prec} & \multicolumn{1}{c}{Rec} & \multicolumn{1}{c}{F} & 
\multicolumn{1}{c}{Time} & \multicolumn{2}{c}{\system{YAGO}$\:\subseteq\:$\system{IMDb}}& \multicolumn{2}{c}{\system{IMDb}$\:\subseteq\:$\system{YAGO}}& 
Time & \multicolumn{2}{c}{\system{YAGO}$\:\subseteq\:$\system{IMDb}}& \multicolumn{2}{c}{\system{IMDb}$\:\subseteq\:$\system{YAGO}}\\
to prev.    & & & &                                &      &
\multicolumn{1}{c}{Num}& \multicolumn{1}{c}{Prec}& \multicolumn{1}{c}{Num}& \multicolumn{1}{c}{Prec}& & \multicolumn{1}{c}{Prec}& \multicolumn{1}{c}{Rec}& \multicolumn{1}{c}{Prec}& \multicolumn{1}{c}{Rec}\\
\midrule
-    & 16h47m & 84\,\% & 75\,\% & 79\,\%       & -  & - & -       & - & -             & 4min  & 91\,\%& 73\,\% & 100\,\% & 60\,\%\\
40.2\,\% & 11h44m & 94\,\% & 89\,\% & 91\,\%       & -  & - & -       & - & -             & 5min  & 91\,\%& 73\,\% & 100\,\% & 80\,\%\\
6.6\,\%  & 11h48m & 94\,\% & 90\,\% & 92\,\%       & -  & - & -       & - & -             & 5min  & 100\,\%& 80\,\% & 100\,\% & 80\,\%\\
0.2\,\%& 11h44m & 94\,\% & 90\,\% & 92\,\%       & 2h17m  & 8 & 100\,\%   & 135k & 28\,\%& 6min  & 100\,\%& 80\,\% & 100\,\% & 80\,\%\\
\bottomrule
\end{tabular}
\caption{Results on matching \system{YAGO} and \system{IMDb} over
iterations 1--4}\label{yires}
\end{table}
The relations are also evaluated manually in both directions. We consider only
the maximally assigned relation, because the relations do not form a hierarchy in \system{YAGO} and \system{DBpedia}. In most cases one assignment dominates
clearly. Table~\ref{ydrel} shows some of the alignments. \paris{} finds
non-trivial alignments of more fine-grained relations to more
coarse-grained ones, of inverses, of symmetric relations, and of relations with completely different names. There are a few plainly wrong alignments, but most errors come from semantic differences that do not show in practice (e.g., \emph{burialPlace} is semantically different from \emph{deathPlace}, so we count it as an error, even though in most cases the two will coincide). Recall is hard to estimate, because not all relations have a counterpart in the other ontology and some relations are poorly populated. We only note that we find alignments for half of \system{YAGO}'s relations in \system{DBpedia}.

\paragraph{\system{YAGO} vs.\ \system{IMDb}}
Next, we were interested in the performance of \paris{} on ontologies
that do not derive from from the same source. For this purpose, we
constructed an RDF ontology from the \system{IMDb}. \system{IMDb} is
predestined for the matching, because it is huge and there is an existing
gold standard: \system{YAGO} contains some mappings to \system{IMDb} movie identifiers, and
we could construct such a mapping for many persons from Wikipedia
infoboxes.
The content of the \system{IMDb} database is available for download as
plain-text files.\footnote{\url{http://www.imdb.com/interfaces\#plain}} The
format of each file is \emph{ad hoc} but we transformed the content of
the database in a fairly straightforward manner into a collection of
triples. For instance, the file \texttt{actors.list} lists for each actor
$x$ the list of all movies $y$ that $x$ was cast in, which we transformed into
facts $\mathit{actedIn(x,y)}$. Unfortunately, the plain-text database
does not contain \system{IMDb} movie and person identifiers (those that
we use for comparing to the gold standard). Consequently, we had to obtain
these identifiers separately. For this purpose, and to avoid having to
access each Web page of the \system{IMDb} Web site, which would require
much too many Web server requests, we used the advanced search feature of
\system{IMDb}\footnote{\url{http://akas.imdb.com/search/}} to obtain
the list of all movies from a given year, or of all persons born in a
certain year, together with their identifiers and everything needed to
connect to the plain-text databases.
Since our \system{IMDb} ontology has only 24 relations, we manually created a gold standard for relations, aligning 15 of them to \system{YAGO} relations.
As Table~\ref{yires} shows, \paris{} took much longer for each iteration
than in the previous experiment. The results are convincing, with an
F-score of 92\,\% for the instances. This is a considerable improvement
over a baseline approach that aligns entities by matching their
\emph{rdfs:label} properties (achieving 97\,\% precision and only 70\,\%
recall, with an F-score of 82\,\%).
Examining by hand the few remaining alignment errors revealed the
following patterns:
\begin{itemize}
\item Some errors were caused by errors in \system{YAGO}, usually caused
by incorrect references from Wikipedia pages to \system{IMDb} movies.
\item \paris{} sometimes aligned instances in \system{YAGO} with
instances in \system{IMDb} that were not equivalent, but very closely
related: for
example,
\emph{King of the Royal Mounted} was aligned with \emph{The
Yukon Patrol}, a feature version of this TV series with the same cast and
crew; \emph{Out 1}, a 13-hour movie, was aligned with \emph{Out 1:
Spectre}, its shortened 4-hour variation.
\item Some errors were caused by the very naive string comparison
approach, that fails to discover, e.g., that \emph{Sugata Sanshir\^o} and
\emph{Sanshiro Sugata} refer to the same movie. It is very likely that
using an improved string comparison technique would further increase
precision and recall of \paris.
\end{itemize}
\paris{}
could align 80\,\% of the relations of \system{IMDb} and \system{YAGO},
with a precision of 100\,\%. 
\paris{} mapped half of the
\system{IMDb} classes correctly to more general or equal
\system{\system{YAGO}} classes (at threshold 0). It performs less well in
the other direction. This is because \system{\system{YAGO}} contains
mostly famous people, many of whom appeared in some movie or documentary
on \system{IMDb}. Thus, \paris{} believes that a class such as
\emph{People from Central Java} is a subclass of \emph{actor}.

As
illustrated here, alignment of instances and relations work very well in
\paris{}, whereas class alignment leaves still some room for improvement.
Overall, the results are very
satisfactory, as
this constitutes, to the best of our knowledge, the first holistic
alignment of instances, relations, and classes on some of the world's
largest ontologies, without any prior knowledge, tuning, or training.

\section{Conclusion}\label{sec:conclusion}

We have presented \paris{}, an algorithm for the automated alignment of RDFS ontologies. Unlike most other approaches, \paris{} computes alignments not only for instances, but also for classes and relations. It does not need training data and it does not require any parameter tuning. \paris{} is based on a probabilistic framework that captures the interplay between schema alignment and instance matching in a natural way, thus providing a holistic solution to the ontology alignment problem. Experiments show that our approach works extremely well in practice.

\paris{} does not use any kind of heuristics on relation names, which allows aligning relations with
completely different names. We conjecture that the name heuristics of more traditional schema-alignment techniques could be factored into the model.

Currently, \paris{} cannot deal with structural heterogeneity. If one ontology models an event by a relation (such as \emph{wonAward}), while the other one models it by an event entity (such as \emph{winningEvent}, with relations \emph{winner}, \emph{award}, \emph{year}), then \paris{} will not be able to find matches. The same applies if one ontology is more fine-grained than the other one (specifying, e.g., cities as birth places instead of countries), or if one ontology treats cities as entities, while the other one refers to them by strings. For future work, we plan to address these types of challenges. We also plan to analyze under which conditions our equations are guaranteed to converge. It would also be interesting to apply \paris{} to more than two ontologies. This would further increase the usefulness of \paris{} for the dream of the Semantic Web.

\bibliographystyle{abbrv}
\bibliography{yago}

\begin{thebibliography}{10}

\bibitem{dbpedia}
S.~Auer, C.~Bizer, G.~Kobilarov, J.~Lehmann, R.~Cyganiak, and Z.~G. Ives.
\newblock {DBpedia: A Nucleus for a Web of Open Data}.
\newblock In {\em Proc.\ ISWC}, 2007.

\bibitem{comapp}
D.~Aumueller, H.-H. Do, S.~Massmann, and E.~Rahm.
\newblock Schema and ontology matching with {COMA++}.
\newblock In {\em Proc.\ SIGMOD}, 2005.

\bibitem{lod2}
C.~Bizer.
\newblock Web of linked data. {A} global public data space on the {W}eb.
\newblock In {\em Proc.\ WebDB}, 2010.

\bibitem{lod}
C.~Bizer, T.~Heath, K.~Idehen, and T.~Berners-Lee.
\newblock Linked data on the {W}eb.
\newblock In {\em Proc.\ WWW}, 2008.

\bibitem{datafusion}
J.~Bleiholder and F.~Naumann.
\newblock Data fusion.
\newblock {\em ACM Computing Surveys}, 41(1), 2008.

\bibitem{sameasbeyond}
L.~Ding, J.~Shinavier, Z.~Shangguan, and D.~L. McGuinness.
\newblock {SameAs} networks and beyond: {A}nalyzing deployment status and
  implications of {owl:sameAs} in linked data.
\newblock In {\em Proc.\ ISWC}, 2010.

\bibitem{linkage-survey}
A.~Elmagarmid, P.~Ipeirotis, and V.~Verykios.
\newblock Duplicate record detection: A survey.
\newblock {\em IEEE TKDE}, 19(1):1--16, 2007.

\bibitem{knowitall}
O.~Etzioni, M.~Cafarella, D.~Downey, S.~Kok, A.-M. Popescu, T.~Shaked,
  S.~Soderland, D.~S. Weld, and A.~Yates.
\newblock Web-scale information extraction in {KnowItAll} (preliminary
  results).
\newblock In {\em Proc.\ WWW}, 2004.

\bibitem{oaei2010}
J.~Euzenat, A.~Ferrara, C.~Meilicke, A.~Nikolov, J.~Pane, F.~Scharffe,
  P.~Shvaiko, and H.~Stuckenschmidt.
\newblock Results of the ontology alignment evaluation initiative 2010.
\newblock In {\em Proc.\ OM}, 2010.

\bibitem{ferrara2008}
A.~Ferrara, D.~Lorusso, and S.~Montanelli.
\newblock Automatic identity recognition in the semantic web.
\newblock In {\em Proc.\ IRSW}, 2008.

\bibitem{glaser2009}
H.~Glaser, A.~Jaffri, and I.~Millard.
\newblock Managing co-reference on the semantic {W}eb.
\newblock In {\em Proc.\ LDOW}, 2009.

\bibitem{gracia2009}
J.~Gracia, M.~d'Aquin, and E.~Mena.
\newblock Large scale integration of senses for the semantic {W}eb.
\newblock In {\em Proc.\ WWW}, 2009.

\bibitem{notsame}
H.~Halpin, P.~Hayes, J.~P. McCusker, D.~McGuinness, and H.~S. Thompson.
\newblock When {owl:sameAs} isn't the same: {A}n analysis of identity in linked
  data.
\newblock In {\em Proc.\ ISWC}, 2010.

\bibitem{Hogan2007}
A.~Hogan.
\newblock Performing object consolidation on the semantic {W}eb data graph.
\newblock In {\em Proc.\ I3}, 2007.

\bibitem{somemore}
A.~Hogan, A.~Polleres, J.~Umbrich, and A.~Zimmermann.
\newblock Some entities are more equal than others: statistical methods to
  consolidate linked data.
\newblock In {\em Proc.\ NeFoRS}, 2010.

\bibitem{www2011eqv}
W.~Hu, J.~Chen, and Y.~Qu.
\newblock A self-training approach for resolving object coreference on the
  semantic {W}eb.
\newblock In {\em Proc.\ WWW}, 2011.

\bibitem{howmatchable}
W.~Hu, J.~Chen, H.~Zhang, and Y.~Qu.
\newblock How matchable are four thousand ontologies on the semantic {W}eb.
\newblock In {\em Proc.\ ESWC}, 2011.

\bibitem{isaac2007}
A.~Isaac, L.~Van Der~Meij, S.~Schlobach, and S.~Wang.
\newblock An empirical study of instance-based ontology matching.
\newblock In {\em Proc.\ ISWC}, 2007.

\bibitem{jean2009}
Y.~R. Jean-Mary, E.~P. Shironoshita, and M.~R. Kabuka.
\newblock Ontology matching with semantic verification.
\newblock {\em J.\ Web Semantics}, 7(3):235--251, 2009.

\bibitem{rimom}
J.~Li, J.~Tang, Y.~Li, and Q.~Luo.
\newblock Rimom: A dynamic multistrategy ontology alignment framework.
\newblock {\em IEEE TKDE}, 21(8):1218--1232, 2009.

\bibitem{cyc}
C.~Matuszek, J.~Cabral, M.~Witbrock, and J.~Deoliveira.
\newblock An introduction to the syntax and content of {Cyc}.
\newblock In {\em Proc. AAAI Spring Symposium}, 2006.

\bibitem{sumo}
I.~Niles and A.~Pease.
\newblock Towards a standard upper ontology.
\newblock In {\em Proc.\ FOIS}, 2001.

\bibitem{noessner2010}
J.~Noessner, M.~Niepert, C.~Meilicke, and H.~Stuckenschmidt.
\newblock Leveraging terminological structure for object reconciliation.
\newblock In {\em Proc.\ ESWC}, 2010.

\bibitem{Ponzetto07}
S.~P. Ponzetto and M.~Strube.
\newblock Deriving a large-scale taxonomy from {W}ikipedia.
\newblock In {\em Proc.\ AAAI}, 2007.

\bibitem{l2r}
F.~Sa\"{\i}s, N.~Pernelle, and M.-C. Rousset.
\newblock {L2R}: {A} logical method for reference reconciliation.
\newblock In {\em Proc.\ AAAI}, 2007.

\bibitem{numprob}
F.~Sa\"{\i}s, N.~Pernelle, and M.-C. Rousset.
\newblock Combining a logical and a numerical method for data reconciliation.
\newblock {\em J.\ Data Semantics}, 12:66--94, 2009.

\bibitem{leila}
F.~M. Suchanek, G.~Ifrim, and G.~Weikum.
\newblock Combining linguistic and statistical analysis to extract relations
  from {W}eb documents.
\newblock In {\em KDD}, 2006.

\bibitem{yago}
F.~M. Suchanek, G.~Kasneci, and G.~Weikum.
\newblock {YAGO}: {A} core of semantic knowledge. {U}nifying {WordNet} and
  {Wikipedia}.
\newblock In {\em Proc.\ WWW}, 2007.

\bibitem{sigma}
G.~Tummarello, R.~Cyganiak, M.~Catasta, S.~Danielczyk, R.~Delbru, and
  S.~Decker.
\newblock Sig.ma: live views on the web of data.
\newblock In {\em Proc.\ WWW}, 2010.

\bibitem{iliads}
O.~Udrea, L.~Getoor, and R.~J. Miller.
\newblock Leveraging data and structure in ontology integration.
\newblock In {\em Proc.\ SIGMOD}, 2007.

\bibitem{silk}
J.~Volz, C.~Bizer, M.~Gaedke, and G.~Kobilarov.
\newblock Discovering and maintaining links on the {W}eb of data.
\newblock In {\em Proc.\ ISWC}, 2009.

\bibitem{wang2008}
S.~Wang, G.~Englebienne, and S.~Schlobach.
\newblock Learning concept mappings from instance similarity.
\newblock In {\em Proc.\ ISWC}, 2008.

\bibitem{rdf}
{Word Wide Web Consortium}.
\newblock {RDF Primer (W3C Recommendation 2004-02-10)}.
\newblock \url{http://www.w3.org/TR/rdf-primer/}, 2004.

\end{thebibliography}

\appendix
\section{Global Functionality}\label{app:globalfun}
There are several design alternatives to define the \emph{global functionality}:
\begin{enumerate}
\item We can count the number of statements and divide it by the number of pairs of statements with the same source:

\[fun(r) = \frac{\# x,y: r(x,y)}{\# x,y,y': r(x,y) \wedge
r(x,y')}\]

This measure is very volatile to single sources that have a large number of targets.
\item We can define functionality as the ratio of the number of first arguments to the number of second arguments:

\[fun(r) = \frac{\# x \exists y: r(x,y)}{\# y \exists x:
r(x,y)}\]

This definition is treacherous: Assume that we have $n$ people and $n$
dishes, and the relationship $\textit{likesDish}(x,y)$. Now, assume that all
people like all dishes. Then $\textit{likesDish}$ should have a low functionality,
because everybody likes $n$ dishes. But the above definition assigns a
functionality of $fun(\textit{likesDish})=\frac{n}{n}=1$. 
\item We can average the local functionalities, as proposed in \cite{somemore}:

\begin{align}\notag
fun(r)& =  avg_{x}\; fun(r,x) =  avg_{x}\Big(\frac{1}{\#y:
r(x,y)}\Big)\\\notag &= \frac{1}{\#x \exists y: r(x,y)} \sum_x \frac{1}{\#y:
r(x,y)}\end{align}

However, the local functionalities are ratios, so that the arithmetic
mean is less appropriate.
\item We can average the local functionalities not by the arithmetic mean, but by the harmonic mean instead

\begin{align}\notag fun(r)&= HM_{x} fun(r,x) =  HM_{x}\Big(\frac{1}{\#y:
r(x,y)}\Big)\\\notag &= \frac{\#x  \exists y: r(x,y)}{\sum_x \#y: r(x,y)} = \frac{\#x
 \exists y: r(x,y)}{\#x,y: r(x,y)}.
\end{align}

\item We may say that the global functionality is the number of first arguments per relationship instance:

\[fun(r)= \frac{\#x  \exists y: r(x,y)}{\#x,y: r(x,y)}\]

This notion is equivalent to the harmonic mean.
\end{enumerate}
With these considerations in mind, we chose the harmonic mean for the definition of the global functionality.

\section{Probabilistic Modeling of First-Order Formulas}\label{app:problog}
In Section~\ref{sec:model}, we presented our probabilistic model of
ontology alignment based on descriptions as first-order sentences of 
our rules, such as Equation~\eqref{eq:punishment}, reproduced here:

\[
\exists r,y: r(x,y) \wedge (\forall y' :r(x',y') \Rightarrow
y {\not\equiv} y')
\wedge fun(r) \mbox{ is high } \Longrightarrow x {\not\equiv}
x
\]

\noindent We derive from these equations probability assessments, such as
Equation~\eqref{p2}, by assuming mutual independence of all
distinct elements of our models (instance equivalence, functionality,
relationship inclusion, etc.). This assumption is of
course not true in practice but it allows us to approximate efficiently the
probability of the consequence of our alignment rules in a canonical manner.
Independence allows us to use the following standard identities:
\begin{align}\notag
P(A\land B)&=P(A)\times P(B)\\\notag
P(A\lor B)&=1-(1-P(A))(1-P(B))\\\notag
P(\forall x: \phi(x))&=\prod_x P(\phi(x))\\\notag
P(\exists x: \phi(x))&=1 -\prod_x (1-P(\phi(x)))\\\notag
\E(\#x: \phi(x))&=\sum_{x} P(\phi(x))
\end{align}
Then, a rule $\phi\Longrightarrow\psi$
is translated as a probability assignment
$P(\psi) := P(\phi)$ and $\phi$ is recursively decomposed using these
identities. Following the example of Equation~\eqref{eq:punishment}, we
derive the value of $\P_2(x\equiv x')$ in Equation~\eqref{p2} as follows:
\begin{eqnarray}\notag
&1-P(\exists r,y\, r(x,y) \wedge (\forall y'\,  r(x',y')\Rightarrow
y\not\equiv y')\wedge fun(r) \mbox{ is high})\\
={}&\prod_{r,y}(1-P(r(x,y))\times\prod_{y'}(1-P(r(x',y')\land
y\equiv y')\times fun(r)
)\\\notag
={}&\prod_{r(x,y)}(1-fun(r)\prod_{r(x',y')}(1-P(y\equiv
y'))).
\end{eqnarray}
since $r(x,y)$ and $r(x',y')$ are crisp, non-probabilistic facts.

Similarly, when we need to estimate a number such as ``$\#
x:\phi(x)$'',
we compute $\mathbb{E}(\#x:\phi(x))$ using the aforementioned
identities.
 
\section{Equivalence of Sets}\label{app:eqvalt}
We compare two instances for equivalence by comparing every statement
about the first instance with every statement about the second instance
(if they have the same relation). This entails a quadratic number of
comparisons. For example, if an actor~$x$ acted in the movies $y_1$,
$y_2$, $y_3$, and an actor~$x'$ acted in the movies $y_1'$, $y_2'$, $y_3'$, then we will compare every statement \emph{actedIn}$(x,y_i)$ with every statement \emph{actedIn}$(x',y_j')$. Alternatively, one could think of the target values as a set and of the relation as a function, as in \emph{actedIn}$(x,\{y_1, y_2, y_3\})$ and \emph{actedIn}$(x',\{y_1', y_2', y_3'\})$. Then, one would have to compare only two sets instead of a quadratic number of statements. However, all elements of one set are potentially equivalent to all elements of the other set. Thus, one would still need a quadratic number of comparisons. 

One could generalize a set equivalence measure (such as the Jaccard
index) to sets with probabilistic equivalences. However, one would still
need to take into account the functionality of the relations: If two
people share an e-mail address (high inverse functionality), they are
almost certainly equivalent. By contrast, if two people share the city
they live in, they are not necessarily equivalent. To unify two
instances, it is sufficient that they share the value of one highly
inverse functional relation. Conversely, if two people have a different birth date, they are certainly different. By contrast, if they like two different books, they could still be equivalent (and like both books). Our model takes this into account. Thus, our formulas can be seen as a comparison measure for sets with probabilistic equivalences, which takes into account the functionalities. 

\end{document}